\pdfoutput=1

\documentclass[11pt]{article}

\usepackage{ACL2023}
\usepackage{times}
\usepackage{latexsym}

\usepackage[T1]{fontenc}

\usepackage[utf8]{inputenc}

\usepackage{microtype}

\usepackage{inconsolata}

\usepackage{wrapfig}
\usepackage{times}
\usepackage{latexsym}

\definecolor{darkblue}{rgb}{0.0,0.0,0.5}

\usepackage{amsmath}
\usepackage{array}
\newcolumntype{L}{>{$}l<{$}}
\newcolumntype{C}{>{$}c<{$}}
\newcolumntype{R}{>{$}r<{$}}

\usepackage{multirow}
\usepackage{tabularx, caption, boldline}
\usepackage{graphicx}
\usepackage{cellspace}
\usepackage{algpseudocode}  
\usepackage{multicol}  
\usepackage{flexisym}

\usepackage{microtype}

\usepackage{dsfont}

\makeatletter
\def\hlinewd#1{%
\noalign{\ifnum0=`}\fi\hrule \@height #1 %
\futurelet\reserved@a\@xhline}
\makeatother

\usepackage{times}
\usepackage{latexsym}

\usepackage{algpseudocode}  
\usepackage{amsmath}
\usepackage{multicol}  
\usepackage{multirow} 
\usepackage{graphicx}  
\usepackage{array}
\usepackage{arydshln}
\usepackage{booktabs}
\usepackage{textcomp}

\usepackage{amssymb}
\usepackage{pifont}

\usepackage{cleveref}
\crefformat{section}{\S#2#1#3} 
\crefformat{subsection}{\S#2#1#3}
\crefformat{subsubsection}{\S#2#1#3}

\usepackage{microtype}

\usepackage{adjustbox}
\usepackage{multicol}  
\usepackage{listings}
\usepackage{multirow} 
\usepackage{amsmath}
\usepackage{amsfonts}
\usepackage{makecell}
\usepackage{algpseudocode} 
\usepackage{verbatim}
\usepackage{mathtools}
\DeclareMathOperator*{\argmax}{arg\,max}

\usepackage{booktabs}

\usepackage{makecell}
\usepackage{cleveref}
\usepackage[normalem]{ulem}

\usepackage{times}
\usepackage{latexsym}

\usepackage{footnote}
\makesavenoteenv{tabular}
\makesavenoteenv{table}

\usepackage[hang,flushmargin]{footmisc}

\usepackage[linesnumbered,ruled]{algorithm2e}
\SetAlFnt{\small}
\SetAlCapNameFnt{\normalsize}
\makeatletter
\newcommand{\nosemic}{\renewcommand{\@endalgocfline}{\relax}}
\newcommand{\dosemic}{\renewcommand{\@endalgocfline}{\algocf@endline}}
\let\oldnl\nl
\newcommand{\nonl}{\renewcommand{\nl}{\let\nl\oldnl}}
\makeatother

\usepackage{enumitem}
\usepackage{tablefootnote}
\usepackage{dblfloatfix}
\usepackage{transparent}
\usepackage{dsfont}

\usepackage{etoolbox}
\makeatletter
\AfterEndEnvironment{algorithm}{\let\@algcomment\relax}
\AtEndEnvironment{algorithm}{\kern2pt\hrule\relax\vskip3pt\@algcomment}
\let\@algcomment\relax
\newcommand\algcomment[1]{\def\@algcomment{\footnotesize#1}}

\usepackage{tabularx}
\makeatletter
\def\hlinewd#1{%
\noalign{\ifnum0=`}\fi\hrule \@height #1 %
\futurelet\reserved@a\@xhline}
\makeatother

\usepackage{minitoc}

\usepackage{cleveref}
\crefformat{section}{\S#2#1#3}
\crefformat{subsection}{\S#2#1#3}
\crefformat{subsubsection}{\S#2#1#3}
\crefrangeformat{section}{\S\S#3#1#4 to~#5#2#6}
\crefmultiformat{section}{\S\S#2#1#3}{ and~#2#1#3}{, #2#1#3}{ and~#2#1#3}

\title{Momentum Decoding: Open-ended Text Generation As Graph Exploration}

\author{
 \textbf{Tian Lan}$^{\heartsuit,}\thanks{~~The first two authors contributed equally.}$ \quad
 \textbf{Yixuan Su}\footnotemark[1]  \quad
  \textbf{Shuhang Liu}$^\heartsuit$ \quad
 \textbf{Heyan Huang}$^{\heartsuit,\clubsuit}$  \quad
 \textbf{Xian-Ling Mao}$^{\heartsuit,}$\thanks{~~Corresponding author.}  \\
 $^\heartsuit$School of Computer Science and Technology, Beijing Institute of Technology\\
 $^\clubsuit$Beijing Engineering Research Center of High Volume Language Information Processing\\
 and Cloud Computing Applications\\
 \texttt{lantiangmftby@gmail.com,\{liush,hhy63,maoxl\}@bit.edu.cn}
}

\begin{document}
\maketitle
\begin{abstract}
Open-ended text generation with autoregressive language models (LMs) is one of the core tasks in natural language processing. However, maximization-based decoding methods (e.g., greedy/beam search) often lead to the degeneration problem, i.e., the generated text is unnatural and contains undesirable repetitions. Existing solutions to this problem either introduce randomness prone to incoherence or require a look-ahead mechanism that demands extra computational overhead. 
In this study, we formulate open-ended text generation from a new perspective, i.e., we view it as an exploration process within a directed graph. Thereby, we understand the phenomenon of degeneration as circular loops within the directed graph. Based on our formulation, we propose a novel decoding method---\textit{momentum decoding}---which encourages the LM to \textit{greedily} explore new nodes outside the current graph. Meanwhile, it also allows the LM to return to the existing nodes with a momentum downgraded by a pre-defined resistance function. We extensively test our approach on three benchmarks from different domains through automatic and human evaluations. The results show that momentum decoding performs comparably with the current state of the art while enjoying notably improved inference speed and computation FLOPs. Furthermore, we conduct a detailed analysis to reveal the merits and inner workings of our approach.\footnote{Our codes and other related resources are publicly available at \url{https://github.com/gmftbyGMFTBY/MomentumDecoding}.}



\end{abstract}

\section{Introduction}
\label{sec:introduction}

Open-ended text generation with autoregressive language models (LMs) is indispensable in various NLP applications. Typical examples include dialogue systems \cite{Thoppilan2022LaMDALM,Su2021PROTOTYPETOSTYLEDG,Rae2021ScalingLM,Su2022MultiTaskPF,su2021dialogue}, contextual text completion \cite{Su2022ACF,Radford2019LanguageMA}, story generation \cite{Mostafazadeh2016ACA,su2022language}, etc.

Conventional maximization-based methods for this task, such as greedy search and beam search, often lead to the degeneration problem~\cite{Holtzman2020TheCC}, i.e., the generated text is unnatural and contains undesirable repetitions.
Existing solutions for this problem can be divided into two categories: 
(1) Stochastic methods, e.g. top-$k$~\cite{fan2018hierarchical} and nucleus sampling~\cite{Holtzman2020TheCC}, introduce randomness to avoid undesirable repetitions. However, the intrinsic stochasticity of these sampling approaches often leads to semantic incoherence and topic drift in the generated text~\cite{basu2020mirostat}. 
(2) Deteriminstic method, i.e., contrastive search~\citep{Su2022ACF,su2022contrastiveiswhatyouneed}, relies on a one-step look-ahead mechanism to encourage diverse generations. While obtaining superior performances, such look-ahead operation demands extra computational overhead.

\begin{figure}[t]     
  \center{\includegraphics[width=0.45\textwidth] {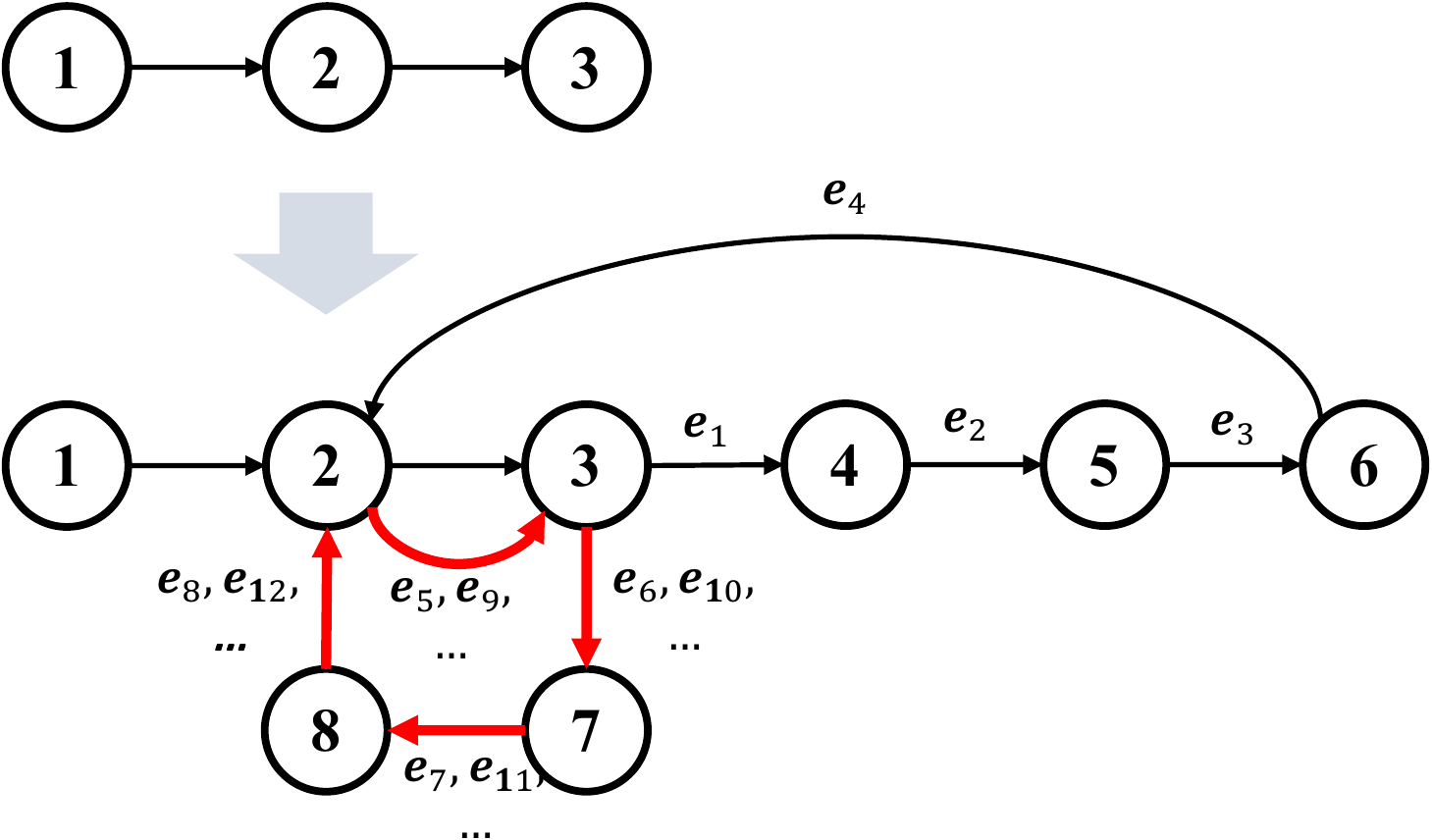}}        
  \caption{An example of the exploration process in the directed graph, where the prefix contains three tokens $[1,2,3]$. $e_i$ denotes the $i$-th decoding step of the LM. The patterns of repetition/degeneration, i.e., $[2,3,7,8]$, are highlighted with red arrows. 
  }
  \label{img:overview}
\end{figure}

In this study, we perceive open-ended text generation from a new perspective. Specifically, we view it as an exploration process within a directed graph.
Therefore, it allows us to formulate the phenomenon of degeneration as circular loops within the directed graph. In Figure~\ref{img:overview}, we provide an illustration in which the LM generates text given a prefix of three tokens, i.e., $[1,2,3]$, and gets stuck in the circular loops, i.e., repetitions, of $[2,3,7,8]$. Intuitively, such degeneration can be addressed if the tendency of the LM to stay in the circular loop can be \textit{properly} discouraged, therefore allowing the LM to jump out of the loop at the correct position and produce text with \textit{natural} repetitions. Based on this motivation, we propose a novel decoding method---\textit{momentum decoding}---which encourages the LM to greedily explore new nodes outside the current graph. Meanwhile, it also allows the LM to return to the existing nodes with a momentum downgraded by a pre-defined resistance function.


Compared with previous methods, we highlight one notable advantage of our proposed approach. 
Specifically, it better bridges the gap between the training and the decoding of the LM. Typically, LMs are trained with the maximum likelihood estimation (MLE) objective. Thereby, at the decoding stage, the LM should follow the same objective~\cite{zhang2019bridging} and try to maximize the likelihood, i.e., probability, of the generated text. However, simply maximizing the likelihood of generation often leads to degeneration. Thus, previous solutions, e.g., sampling and contrastive search, propose to modify the decoding objective at \textit{every} generation step.
In contrast to previous approaches, our proposed method largely follows the greedy objective during decoding. It only corrects the generation at the positions where the symptom of degeneration is clear, e.g., within the circular loops of $[2,3,7,8]$ in Figure~\ref{img:overview}. In the experiments (\cref{sec:automatic_evaluation}), we show that momentum decoding generates text by following the greedy objective for more than 70\% of the decoding steps.

We comprehensively test our approach on three benchmarks from different domains. The automatic evaluations (\cref{sec:automatic_evaluation}) verify that momentum decoding generates the most diverse outputs while maintaining a high semantic coherence in the generated text. Moreover, extensive human evaluations (\cref{sec:human_evaluation}) demonstrate that momentum decoding performs on par with the current state of the art, i.e., contrastive search, but with 30\% of inference speedup and more than 4$\times$ reduction in computation FLOPs. Lastly, we provide in-depth analyses of the inner workings of our approach (\cref{sec:analysis}).

In summary, our contributions are:
\begin{itemize}
    \item A new perspective for understanding the task of open-ended text generation.
    \item The proposal of a novel decoding method---momentum decoding---for generative LMs. 
    \item Extensive experiments and in-depth analyses reveal the proposed method's merits and advantages.
\end{itemize}


\section{Preliminaries}

In this work, we study the fundamental technique, i.e., the autoregressive decoding methods, for open-ended text generation. The autoregressive decoding method repeatedly predicts and selects the next token $x_t$ conditioned on the previous context $\boldsymbol{x}_{<t}$. 
So far, there are two types of methods used for autoregressive decoding, which are (1) deterministic methods and (2) stochastic methods.

\paragraph{Deterministic Methods.}
Two widely used deterministic approaches are greedy and beam search, which aim to select the text continuation with the highest probability based on the model’s probability distribution. However, solely maximizing the output probability often leads to dullness \cite{Li2016ADO} and degeneration \cite{Fan2018HierarchicalNS,Holtzman2020TheCC} in the generated text.

To address this problem, contrastive search~\citep{Su2022ACF} is recently introduced, which proposes a one-step look-ahead mechanism to select the diverse tokens, obtaining the new state of the art on various open-ended text generation benchmarks.

\paragraph{Stochastic Methods.}
To tackle the degeneration problem, stochastic approaches have been proposed to sample the next token from the probability distribution $p_{\theta}(x_t|\boldsymbol{x}_{<t})$. To avoid sampling from the unreliable tail of distribution, \citeauthor{fan2018hierarchical} proposed top-k sampling which draws sample from the vocabulary subset $V^{(k)}$
that maximizes $\sum_{v\in V^{(k)}} p_{\theta}(x_t|\boldsymbol{x}_{<t})$. Here, $|V^{(k)}|= k$ and $\boldsymbol{x}_{<t}$ is the prefix context. Differently, the current state-of-the-art nucleus sampling \cite{Holtzman2020TheCC} draws sample from the smallest vocabulary subset $U$ with total probability mass above a threshold $p\in [0, 1]$; i.e., $U$ is the smallest vocabulary subset such that $\sum_{v\in U} p_{\theta}(x_t|\boldsymbol{x}_{<t}) \geq p$. 
While the sampling approaches help to alleviate model degeneration, 
the intrinsic stochasticity in these methods often leads to 
the semantic incoherence and topic drift \cite{Basu2021MirostatAN}.

\section{Methodology}
In this section, we first formulate open-ended text generation as an exploration process within the directed graph (\cref{sec:formulation}). Then, we understand the phenomenon of degeneration as circular loops within the directed graph (\cref{sec:degeneration_in_graph}). Lastly, based on our motivation, we introduce our solution---momentum decoding---to the degeneration problem (\cref{sec:momentum_decoding}).

\subsection{Open-ended Text Generation as Graph Exploration}
\label{sec:formulation}

We formulate the task of open-ended text generation from a new perspective, i.e., we view it as an exploration process within a directed graph. As an example in Figure~\ref{img:overview}, at the start, the prefix text (i.e., $[1,2,3]$) can be viewed as a directed graph $G$, which consists of three nodes and two directed edges. Then, throughout the generation process of the LM, more nodes are included in the directed graph $G$, e.g., the generated tokens $[4,5,6,7,8]$. We note that the nodes within the graph could be accessed multiple times during the generation of the LM. For instance, these nodes might correspond to frequently mentioned name entities, e.g., the tokens $[2,3]$ in Figure~\ref{img:overview}.



\subsection{The Phenomenon of Degeneration}
\label{sec:degeneration_in_graph}
Based on our formulation in \cref{sec:formulation}, we can understand the phenomenon of degeneration as \textit{circular loops} within the directed graph. For instance, in Figure~\ref{img:overview}, the LM gets stuck in the circular loop of $[2,3,7,8]$, which corresponds to the degeneration (i.e., invalid repetition). Then, we define the \textit{circular depth} of a node (i.e., token) based on the current graph (i.e., text sequence). Given the context text $\boldsymbol{x}$ with length $n$ (i.e. $|\boldsymbol{x}|=n$) and the new token $t$, the circular depth of $t$ with respect to $\boldsymbol{x}$ is 

\begin{equation}
    \label{eq:depth_function}
  d(t;\boldsymbol{x}) =
    \begin{cases}
      0 & \text{if $t\notin \boldsymbol{x}$}\\
      \max_{1\leq i\leq n+1, \textup{s.t.} [\boldsymbol{x}_{(i\rightarrow n)}:t]\in\boldsymbol{x}}\{&\\
      \|[\boldsymbol{x}_{(i\rightarrow n)}:t]\|\}& \text{if $t\in \boldsymbol{x}$}\\
    \end{cases}       
\end{equation}
where $\boldsymbol{x}_{(i\rightarrow n)}$ denotes the subsequence of $\boldsymbol{x}$ from its $i$-th token to its $n$-th token, and $\boldsymbol{x}_{(n+1\rightarrow n)}$ is an empty string $\phi$. The 
$[:]$ denotes the concatenation operation and $\|\cdot\|$ is the length of the text sequence.

\subsection{Momentum Decoding}
\label{sec:momentum_decoding}
We propose a novel decoding method---\textit{momentum decoding}. The basic principle behind momentum decoding (MD) is straightforward. During generation, MD encourages the LM to \textit{greedily} explore new nodes outside the current graph. Meanwhile, it also allows the LM to return to the existing nodes with a momentum downgraded by a pre-defined resistance function. Our rationale is to prevent the LM from generating \textit{deep} circular loops as such loops often lead to severe degenerations~\citep{su2022contrastiveiswhatyouneed}. 

Formally, given the prefix text $\boldsymbol{x}_{<t}$, the LM first looks at the most probable token $\hat{v}$. Here, $\hat{v}=\argmax_{v\in\mathcal{V}}p_{\theta}(v|\boldsymbol{x}_{<t})$, where $\mathcal{V}$ is the LM's vocabulary and $p_{\theta}(\cdot|\boldsymbol{x}_{<t})$ is the probability distribution produced by the LM. Then, the selection of the output token $x_t$ follows
\begin{equation}
    \label{eq:momentum_decoding}
  x_t =
    \begin{cases}
      \hat{v} & \text{if $\hat{v}\notin \boldsymbol{x}_{<t}$}\\
      \argmax_{c\in \mathcal{C}^{(k)}}\bigg\{p_{\theta}(c|\boldsymbol{x}_{<t}) & \\
      \quad\quad\quad -  \alpha\times f(d(c;\boldsymbol{x}_{<t}))\bigg\}& \text{if $\hat{v}\in \boldsymbol{x}_{<t}$}\\
    \end{cases}       
\end{equation}
where $\mathcal{C}^{(k)}$ is the set of top-$k$ predictions from the LM's probability distribution $p_{\theta}(\cdot|\boldsymbol{x}_{<t})$, and $k$ is typically set as $5$. The function $f(\cdot)$ is a pre-defined resistance function, and $d(\cdot)$ is defined in Eq. (\ref{eq:depth_function}). And $\alpha$ is a hyperparameter that regulates the importance of the two components.

\begin{table}[t]
    \small
	\centering  
	\renewcommand{\arraystretch}{1.2}
	\setlength{\tabcolsep}{6pt}
	\scalebox{1.1}{
	\begin{tabular}{cc}
	    \hlinewd{0.75pt}
            Circular Depth $d(\cdot)$&Output Resistance $f(d(\cdot))$\\
            \hline
            1&1.0\\
            2&3.0\\
            3&4.0\\
            $\geq$4&5.0\\
		\hlinewd{0.75pt}
	\end{tabular}}
    \caption{Look-up table for the resistance function.}
    	\vspace{-1.5mm}
	\label{tb:resistance_function}
\end{table}

In this work, we design a conceptually simple yet empirically effective form of $f(\cdot)$. Specifically, $f(\cdot)$ is defined as a look-up table\footnote{We acknowledge that the design of $f(\cdot)$ is very flexible. This study uses a look-up table for its empirical simplicity and computational efficiency. We leave the more sophisticated design of $f(\cdot)$ to our future work.} as shown in Table~\ref{tb:resistance_function}. Intuitively, when the candidate $c$ leads to a circular loop in the existing graph, the deeper the loop is, the more resistance it will receive from the resistance function. Thereby, the LM is encouraged to jump out of the loop and explore new nodes outside the current graph. 

In Algorithm~\ref{alg:momentum_decoding}, we illustrate the decoding process of momentum decoding.

\normalem
\begin{algorithm}[t]
   \SetAlCapFnt{\small}
   \SetAlCapNameFnt{\small}
    \caption{Momentum Decoding}
    \SetKwInOut{Input}{Input}
    \SetKwInOut{Output}{Output}
    \label{alg:momentum_decoding}
    
    \Input{The LM $ \theta$ (e.g. GPT-2); the vocabulary of the LM $\mathcal{V}$; the prefix text $\boldsymbol{x}$; the maximum generation step $T$.}
    Initialize the directed graph $G$ with the prefix text $\boldsymbol{x}$;\\
    \For{step $t=1, ..., T$}
    {
        Compute the next token probability $p_{\theta}(\cdot|\boldsymbol{x})$;\\
        Get the most probable token $\hat{v} $ as $\hat{v}=\argmax_{v\in\mathcal{V}}p_{\theta}(v|\boldsymbol{x})$;
        
        
        \eIf{$\hat{v} \notin G$}{
            $\hat{x}=\hat{v}$;\\
        }{
            Collect the set of top-$k$ candidate tokens $\mathcal{C}^{(k)}$ from $p_{\theta}(\cdot|\boldsymbol{x})$;\\
            $\hat{x}=\argmax_{c\in \mathcal{C}^{(k)}}\bigg\{p_{\theta}(c|\boldsymbol{x}_{<t}) - \alpha\times f(d(c;\boldsymbol{x}))\bigg\}$; (see Eq. (\ref{eq:momentum_decoding}))\\
        }
        Update the directed graph $G$ with next token $\hat{x}$;\\
        Update the prefix $\boldsymbol{x}=[\boldsymbol{x}:\hat{x}]$;
    }
    \Output{The generated text $\boldsymbol{x}$.}
\end{algorithm}

\section{Experiments}
In this section, we provide details of our experimental setups and evaluation results.

\paragraph{Evaluation Benchmarks.} 
Following previous studies \cite{su2022empirical,Li2022ContrastiveDO}, we conduct extensive experiments on three open-ended text generation benchmarks from different domains, including (i) articles from Wikinews\footnote{\url{https://www.wikinews.org}} in the news domain; (ii) Wikitext-103 dataset \cite{Merity2017PointerSM} from the Wikipedia domain; (iii) and BookCorpus \cite{Zhu2015AligningBA} from the story domain.

\paragraph{Model and Baselines.} We compare momentum decoding with a range of existing decoding methods, including (1) greedy search (Greedy); (2) beam search (Beam); (3) top-$k$ sampling (Top-$k$)~\citep{fan2018hierarchical}; (3) nucleus sampling (Nucleus)~\citep{Holtzman2020TheCC}; (4) typical sampling (Typical)~\citep{Meister2022TypicalDF}; (5) contrastive decoding (CD)~\citep{Li2022ContrastiveDO};\footnote{During generation, contrastive decoding demands an extra amateur LM. In our experiments, we follow~\citet{Li2022ContrastiveDO} and use the GPT-2-small model as the amateur LM.}  and (6) contrastive search (CS)~\cite{Su2022ACF}. For the proposed momentum decoding (MD), we set the $k$ and $\alpha$ (see Eq. (\ref{eq:momentum_decoding})) as $5$ and $0.2$, respectively.\footnote{We use the hyperparameters of different baseline methods as suggested by previous studies~\citep{Li2022ContrastiveDO,Su2022ACF}. The hyperparameters of MD are selected based on the LM's performance on the validation set.}

Following~\citet{su2022empirical,Li2022ContrastiveDO}, we use the GPT-2-XL model~\citep{Radford2019LanguageMA} as the evaluated LM. The generation of the LM is conditioned on the test prompts with a fixed length of 32. And the generation of the text ends upon reaching an end-of-document token or a maximum length of 256 tokens.

\begin{table*}[t]
    \small
	\centering  
	\renewcommand{\arraystretch}{1.2}
	\setlength{\tabcolsep}{6pt}
	\scalebox{1.0}{
	\begin{tabular}{cccccccr}
	    \hlinewd{0.75pt}
            &\textbf{Method}&Diversity(\%)$\uparrow$&MAUVE(\%)$\uparrow$&Coherence$\uparrow$&Greedy Ratio(\%)$\uparrow$&MD-Speedup&FLOPs$\downarrow$\\
            \cline{2-8}
            \multirow{8}{*}{\rotatebox[origin=c]{90}{{\textbf{News}}}}&Greedy&3.55&13.89&-0.47&\textbf{100.00}&$\Delta0\%$&1.0$\times$\\
            &Beam&5.62&8.04&\textbf{-0.45}&90.04&$\Delta36\%$&4.0$\times$\\
            &Top-$k$&91.56&89.41&-2.22&52.95&$\Delta2\%$&1.0$\times$\\
            &Nucleus&93.54&88.86&-2.61&48.59&$\Delta0\%$&1.0$\times$\\
            &Typical&91.21&90.80&-2.02&52.95&$\Delta7\%$&1.0$\times$\\
            &CD&92.61&\textbf{92.90}&-2.27&35.58&$\Delta38\%$&5.0$\times$\\
            &CS&93.72&80.87&-1.39&72.80&$\Delta31\%$&4.36$\times$\\
            &MD(ours)&\textbf{97.66}&76.93&-1.34&77.95&-&1.0$\times$\\
		\hlinewd{0.75pt}
            &\textbf{Method}&Diversity(\%)$\uparrow$&MAUVE(\%)$\uparrow$&Coherence$\uparrow$&Greedy Ratio(\%)$\uparrow$&MD-Speedup&FLOPs$\downarrow$\\
            \cline{2-8}
            \multirow{8}{*}{\rotatebox[origin=c]{90}{{\textbf{Wikipedia}}}}&Greedy&3.40&6.02&-0.41&\textbf{100.00}&$\Delta0\%$&1.0$\times$\\
            &Beam&2.93&3.82&\textbf{-0.40}&91.90&$\Delta18\%$&4.0$\times$\\
            &Top-$k$&90.33&84.89&-2.37&50.80&$\Delta5\%$&1.0$\times$\\
            &Nucleus&94.25&\textbf{91.57}&-3.03&43.55&$\Delta5\%$&1.0$\times$\\
            &Typical&86.89&85.24&-2.21&50.80&$\Delta8\%$&1.0$\times$\\
            &CD&90.73&90.78&-2.34&36.38&$\Delta38\%$&5.0$\times$\\
            &CS&89.82&79.52&-1.56&67.60&$\Delta25\%$&4.36$\times$\\
            &MD(ours)&\textbf{97.12}&83.94&-1.55&74.37&-&1.0$\times$\\
		\hlinewd{0.75pt}
            &\textbf{Method}&Diversity(\%)$\uparrow$&MAUVE(\%)$\uparrow$&Coherence$\uparrow$&Greedy Ratio(\%)$\uparrow$&MD-Speedup&FLOPs$\downarrow$\\
            \cline{2-8}
            \multirow{8}{*}{\rotatebox[origin=c]{90}{{\textbf{Story}}}}&Greedy&0.86&2.67&-0.34&\textbf{100.00}&$\Delta0\%$&1.0$\times$\\
            &Beam&1.44&2.0&\textbf{-0.32}&93.07&$\Delta36\%$&4.0$\times$\\
            &Top-$k$&91.22&86.38&-2.45&45.03&$\Delta2\%$&1.0$\times$\\
            &Nucleus&94.50&\textbf{91.77}&-3.02&41.56&$\Delta0\%$&1.0$\times$\\
            &Typical&90.41&85.77&-2.26&47.03&$\Delta7\%$&1.0$\times$\\
            &CD&89.66&91.13&-2.23&35.63&$\Delta35\%$&5.0$\times$\\
            &CS&93.06&51.82&-1.61&69.54&$\Delta28\%$&4.36$\times$\\
            &MD(ours)&\textbf{96.99}&67.67&-1.47&73.86&-&1.0$\times$\\
		\hlinewd{0.75pt}
	\end{tabular}}
    \caption{Automatic evaluation results. $\uparrow$ means the higher the better and $\downarrow$ means the lower the better.}
    	\vspace{-1.5mm}
	\label{tb:main_results}
\end{table*}

\subsection{Automatic Evaluation}
\label{sec:automatic_evaluation}
\subsubsection{Evaluation Metrics}
We follow previous studies~\citep{Li2022ContrastiveDO,Su2022ACF,su2022contrastiveiswhatyouneed} and use the metrics below for automatic evaluation.

\noindent(1) \textbf{Diversity} takes into account the generated repetition at different $n$-gram levels and it is defined as: $\textup{{diversity}}=\prod_{n=2}^{4}(1.0-\frac{\textup{rep-n}}{100})$, where $\textup{{rep-n}}=100 \times (1.0 - \frac{|\textup{unique n-grams}(\hat{\boldsymbol{x}})|}{|\textup{total n-grams}(\hat{\boldsymbol{x}})|})$ and $\hat{\boldsymbol{x}}$ is the text generated by the LM.

\noindent(2) \textbf{MAUVE}~\citep{Pillutla2021MAUVEMT} is a metric designed for measuring the token distribution closeness between the generated text and human-written text. However, as recently pointed out by~\citet{su2022empirical}, MAUVE does not accurately reflect human preferences over different decoding methods.

\noindent(3) \textbf{Coherence}~\citep{su2022contrastiveiswhatyouneed} automatically measures the semantic coherence between the prefix 
text $\boldsymbol{x}$ and the generated text $\hat{\boldsymbol{x}}$ using a massively pre-trained OPT-2.7B LM~\cite{Zhang2022OPTOP}.
Specifically, the metric is defined as the averaged log-likelihood of the generated text conditioned on the prefix text as:
\begin{equation}
    \begin{split}
        \frac{1}{|\hat{\boldsymbol{x}}|} \sum_{i=1}^{|\hat{\boldsymbol{x}}|} \log p_{\mathcal{M}}\left(\hat{\boldsymbol{x}}_i \mid\left[\boldsymbol{x}: \hat{\boldsymbol{x}}_{<i}\right]\right),
    \end{split}
    \label{eq:coherence}
\end{equation}
where $[:]$ is the concatenation operation.

\noindent(4) \textbf{Greedy Ratio} measures the proportion of the LM's generation equal to the prediction of the maximization-based method, i.e., greedy search. Formally, given the prefix $\boldsymbol{x}$ and the output $\hat{\boldsymbol{x}}=\{\hat{x}_1,...,\hat{x}_{|\hat{\boldsymbol{x}}|}\}$ generated by the LM, the greedy ratio is defined as
\begin{equation}
    \frac{1}{|\hat{\boldsymbol{x}}|}\sum_{i=1}^{|\hat{\boldsymbol{x}}|}\mathds{1}(\hat{x}_i=\argmax_{v\in\mathcal{V}}p_{\theta}(v|[\boldsymbol{x}:\hat{\boldsymbol{x}}_{<i}])),
\end{equation}
where $\mathcal{V}$ is the vocabulary of the LM $\theta$, and $[:]$ denotes the concatenation operation. We note that a higher greedy ratio indicates the decoding method behaves more similarly to greedy search. Thereby, it better bridges the gap between the training and the decoding of the LM, as described in \cref{sec:introduction}.

\noindent (5) \textbf{MD-Speedup} computes the relative per token inference speedup of momentum decoding with respect to different compared methods. 

\noindent (6) \textbf{FLOPs} measures the computational complexity of different methods in terms of the number of required floating-point operations during inference. A higher FLOPs means the method is computationally more intensive~\cite{Liu2020FastBERTAS}.\footnote{We use the deepspeed package (\url{https://github.com/microsoft/DeepSpeed}) to calculate the FLOPs of different decoding methods.}

\begin{table*}[t]
    \small
	\centering  
	\renewcommand{\arraystretch}{1.2}
	\setlength{\tabcolsep}{6pt}
	\scalebox{1.0}{
	\begin{tabular}{cccccc}
	    \hlinewd{0.75pt}
	    \multirow{4}{*}{\rotatebox[origin=c]{90}{{\textbf{News}}}}&\multicolumn{2}{c}{Method A is better}&Neutral&\multicolumn{2}{c}{Method B is better}\\
	    \cmidrule(lr){2-3}
	    \cmidrule(lr){4-4}
	    \cmidrule(lr){5-6}
            &Momentum Decoding&\textbf{56.7}\%$^{\dagger}$&1.3\%&42.0\%&Nucleus Sampling\\
            &Momentum Decoding&\textbf{57.0}\%$^{\dagger}$&3.0\%&40.0\%&Contrastive Decoding\\
            &Momentum Decoding&40.7\%&8.0\%&\textbf{51.3}\%$^{\dagger}$&Contrastive Search\\
	    \hlinewd{0.75pt}
	    \multirow{4}{*}{\rotatebox[origin=c]{90}{{\textbf{Wikipedia}}}}&\multicolumn{2}{c}{Method A is better}&Neutral&\multicolumn{2}{c}{Method B is better}\\
	    \cmidrule(lr){2-3}
	    \cmidrule(lr){4-4}
	    \cmidrule(lr){5-6}
            &Momentum Decoding&\textbf{60.0}\%$^{\dagger}$&0.7\%&39.3\%&Nucleus Sampling\\
            &Momentum Decoding&\textbf{62.5}\%$^{\dagger}$&4.5\%&33.0\%&Contrastive Decoding\\
            &Momentum Decoding&\textbf{50.0}\%$^{\parallel}$&7.3\%&42.7\%$^{\parallel}$&Contrastive Search\\
	    \hlinewd{0.75pt}
	    \multirow{4}{*}{\rotatebox[origin=c]{90}{{\textbf{Story}}}}&\multicolumn{2}{c}{Method A is better}&Neutral&\multicolumn{2}{c}{Method B is better}\\
	    \cmidrule(lr){2-3}
	    \cmidrule(lr){4-4}
	    \cmidrule(lr){5-6}
            &Momentum Decoding&\textbf{59.0}\%$^{\dagger}$&1.3\%&39.7\%&Nucleus Sampling\\
            &Momentum Decoding&\textbf{58.0}\%$^{\dagger}$&3.0\%&39.0\%&Contrastive Decoding\\
            &Momentum Decoding&\textbf{46.7}\%$^{\parallel}$&6.6\%&\textbf{46.7}\%$^{\parallel}$&Contrastive Search\\
	    \hlinewd{0.75pt}
	\end{tabular}}
    \caption{Human evaluation results. $^{\dagger}$ means one method performs significantly better than the other as judged by Sign Test with $p$-value < 0.05. $^{\parallel}$ means one system performs comparably with the other with $p$-value > 0.4.}
	\label{tb:human_evaluation}
\end{table*}

\subsubsection{Evaluation Results}

Table \ref{tb:main_results} presents the experimental results of the automatic evaluation, from which we can make the following conclusions:

\noindent(1) compared with previous state-of-the-art works, momentum decoding (MD) achieves the highest diversity on three benchmarks. This observation demonstrates that momentum decoding effectively addresses the degeneration problem by preventing the LMs from generating deep, circular loops.

\noindent(2) MD performs notably better than state-of-the-art baselines on the coherence metric, such as contrastive search and contrastive decoding, 
suggesting it best maintains the semantic consistency between the generated text and the given prefix text and the semantic consistency inner the generated text.
Although greedy search and beam search outperforms MD on the coherence, they suffer the severe degeneration problem because of their over-confidence over probability of LMs.

\noindent(3) compared with state-of-the-art baselines, such as contrastive search and contrastive decoding, MD's greedy ratio is much higher. This observation proves that MD's gap between training and inference is smaller, leading to more reliable and robust performance. Similarly, greedy search and beam search achieves the highest greedy ratio, but their generations face a serious degeneration problem.

\noindent(4) the MAUVE scores of contrastive search and momentum decoding are weaker than stochastic decoding methods. As pointed out by previous studies~\cite{su2022contrastiveiswhatyouneed,su2022empirical}, MAUVE does accurately reflect the actual performance of baselines. For example, nucleus sampling achieves a higher MAUVE score than contrastive search, which contradicts the human evaluation in previous works \cite{su2022contrastiveiswhatyouneed,su2022empirical,Su2022ACF}.
In this paper, we analyze the quality of baselines accurately by conducting the human evaluation in Section \cref{sec:human_evaluation}.

\noindent(5) it is worth noting that momentum decoding achieves comparable efficiency with the most efficient autoregressive decoding method, i.e., the greedy search, 
on MD-Speedup and FLOPs metrics, 
and significantly outperforms the state-of-the-art contrastive search baseline by a large margin. 
For example, the FLOPs of contrastive search are over four times that of MD's FLOPs, indicating its much higher computation burden during online inference.
Meanwhile, compared with greedy search, the FLOPs and MD-Speedup of momentum decoding are $\Delta 0\%$ and $1\times$ on three benchmarks, respectively.

\subsection{Human Evaluation}
\label{sec:human_evaluation}
We also conduct a human evaluation with four native-speaker graders from a third-party grading platform. We randomly select 150 test prompts from the benchmarks across different domains. We compare momentum decoding against nucleus sampling, contrastive decoding, and contrastive search (i.e. the current state of the art) through pairwise comparison. Specifically, for each test prompt, the annotators are given two texts, in random order, that are generated by MD and another compared method. The annotators then decide which one is more likely written by humans considering the following aspects of the generated text:

\begin{itemize}[noitemsep,topsep=1pt]
    \itemsep 0em 
    \item \textbf{Coherence}: Whether the generated text is semantically coherent.
    \item \textbf{Fluency}: Whether the generated text is fluent and easy to understand.
    \item \textbf{Informativeness}: Whether the generated text is diverse and contains interesting content. 
\end{itemize}

Table \ref{tb:human_evaluation} presents the experimental results of the human evaluation.
It can be found that momentum decoding significantly outperforms nucleus sampling and contrastive decoding by a large margin on all three benchmarks. 
Moreover, momentum decoding also achieves a comparable performance with the state-of-the-art contrastive search method as judged by Sign Test.
For example, momentum decoding slightly outperforms contrastive search on the Wikipedia domain.
These observations are impressive due to MD's higher inference efficiency than contrastive search, showing its potential for efficient online inference.

\subsection{Case Study}

This section shows a case from the Wikinews domain to compare our proposed momentum decoding and contrastive search (CS).
As shown in Table \ref{tb:case_examples}, three human annotators consistently agree that the generation of momentum decoding is better than CS's result.
We can make two conclusions based on their comparison:
(1) CS's generation is slightly incoherent with the given prefix. This might be because contrastive search overcorrects the top-$1$ candidate token in some decoding steps;
(2) momentum decoding encourages the LM to greedily explore new tokens (highlighted in \textcolor{red}{red}) outside the directed graph, which is degeneration-free and includes diverse tokens to the generation;
(3) momentum decoding also allows the LM to return to the nodes (highlighted in \textcolor{blue}{blue}) within the directed graph, and resistance effectively avoids the degeneration problem in this case.

\begin{table*}[t]
    \small
	\centering  
	\renewcommand{\arraystretch}{1.2}
	\setlength{\tabcolsep}{6pt}
	\scalebox{0.76}{
	\begin{tabular}{cc}
	    \hlinewd{0.75pt}
        \textbf{Method}&\textbf{Generated Result}\\
        \hline
        \makecell[c]{Contrastive\\  Search}&\makecell[l]{\underline{\textbf{Slovakia eliminated defending champions Italy from the World Cup. First round groups E and F were decided on Thursday: Japan,}}\\ \underline{\textbf{Paraguay and the Netherlands progress alongside}} Slovakia.\\\\In a statement, FIFA said it would take ``immediate action'' in relation to the match-fixing allegations, which came to light last week in a report\\ by the South African newspaper Rapport.\\\\FIFA president Sepp Blatter was due to hold a news conference on Friday in Zurich but postponed the event to allow him to ``consult with legal\\ advisers on the best way to deal with this situation,'' the statement said.\\\\Rapport's report alleged that a group of players, including Italy's Andrea Pirlo, had agreed to fix matches in exchange for cash payments, and\\ that the money was used to pay for Pirlo's birthday present to his wife.\\\\Blatter, who has been in office since 1998, said in a statement: ``I am shocked and dismayed by the news that has come to light in the last few\\ days. It is a sad day for football and the world of sport in general, which is based on fair play and the rule of law.''\\\\The Swiss Football Association (SFSA) said it had opened an investigation into the allegations and was in contact with FIFA's ethics committee.\\\\Slovakia's Prime Minister Robert Fico told public broadcaster SVT ...}\\
        \hline
        \makecell[c]{Momentum\\ Decoding}&\makecell[l]{\underline{\textbf{Slovakia eliminated defending champions Italy from the World Cup. First round groups E and F were decided on Thursday: Japan,}}\\ \underline{\textbf{Paraguay and the Netherlands progress alongside}} Slovakia \textcolor{blue}{.}\\\\
        \textcolor{red}{The second}
        \textcolor{blue}{round}
        \textcolor{red}{of matches}
        \textcolor{red}{will be played} 
        \textcolor{blue}{on} 
        \textcolor{red}{Saturday}\textcolor{blue}{,} 
        \textcolor{red}{with} 
        \textcolor{blue}{the}
        \textcolor{red}{top two teams} 
        \textcolor{blue}{from}
        \textcolor{red}{each group progressing to} 
        \textcolor{blue}{the} 
        \textcolor{red}{quarter-finals}\textcolor{blue}{.} 
        The\\ 
        \textcolor{blue}{quarter-}final 
        \textcolor{red}{draw} 
        \textcolor{blue}{will} 
        \textcolor{red}{take place} 
        \textcolor{blue}{on} 
        \textcolor{red}{Sunday}\textcolor{blue}{.}
        \\\\\
        ``\textcolor{red}{We are very happy that}
        \textcolor{red}{we have qualified for} 
        \textcolor{blue}{the last} 
        \textcolor{red}{eight,'' said} 
        \textcolor{blue}{Slovakia} 
        \textcolor{red}{coach Zdenek Zeman}\textcolor{blue}{.}\textcolor{red}{``It is a great achievement} 
        \textcolor{blue}{for our} 
        \textcolor{red}{team} 
        \textcolor{blue}{and} 
        \textcolor{blue}{I}\\ 
        \textcolor{red}{am sure} 
        it
        \textcolor{blue}{will} 
        give 
        \textcolor{red}{us} 
        \textcolor{blue}{a} 
        \textcolor{red}{lot of confidence going into}
        \textcolor{blue}{the} 
        \textcolor{red}{tournament.''}
        \\\\
        Italy's qualification 
        \textcolor{red}{was secured by} 
        a 
        \textcolor{red}{2-1 win over} 
        Croatia
        \textcolor{red}{in} 
        their 
        \textcolor{red}{opening game}\textcolor{blue}{.} 
        They then 
        \textcolor{red}{beat Sweden 1}-0 
        \textcolor{blue}{in} 
        an 
        \textcolor{red}{entertaining} 
        match at 
        \textcolor{blue}{the}\\ 
        \textcolor{red}{Stadio Olimpico} 
        \textcolor{blue}{in} 
        \textcolor{red}{Rome}\textcolor{blue}{.}\\\\
        \textcolor{red}{Croatia}\textcolor{blue}{,} 
        \textcolor{red}{who finished third} 
        \textcolor{blue}{in} 
        Group \textcolor{red}{A},\textcolor{blue}{will} 
        \textcolor{red}{face} 
        either England 
        \textcolor{red}{or Germany} 
        \textcolor{blue}{in the} 
        semi\textcolor{blue}{-finals} 
        after they lost 
        \textcolor{blue}{to} 
        Spain 
        \textcolor{blue}{and} 
        \textcolor{red}{Portugal respectively}
        \textcolor{blue}{.}\\\\
        Group B features hosts 
        \textcolor{red}{France}\textcolor{blue}{,} 
        Belgium\textcolor{blue}{,} 
        Russia and Ukraine. 
        France 
        \textcolor{blue}{beat Ukraine} 
        \textcolor{red}{3}-2 in \textcolor{blue}{Paris on} 
        Wednesday before
        \textcolor{red}{beating} 
        \textcolor{blue}{Russia 2-3} 
        \textcolor{blue}{in} 
        \textcolor{red}{Kiev}\\ 
        \textcolor{blue}{on} 
        Friday\textcolor{blue}{.}
        \\\\
        \textcolor{red}{Belgium}\textcolor{blue}{,} 
        which 
        \textcolor{blue}{finished} 
        fourth 
        \textcolor{blue}{in Group} \textcolor{red}{B}\textcolor{blue}{, will} 
        play 
        \textcolor{blue}{either} 
        Germany 
        \textcolor{blue}{or England in the} final\textcolor{blue}{.}\\\\
        Russia\textcolor{blue}{,} 
        whose qualifying 
        \textcolor{red}{campaign} 
        ended 
        \textcolor{blue}{in} 
        disappointment\textcolor{blue}{, will} 
        meet 
        \textcolor{blue}{either} 
        Spain
        \textcolor{blue}{or Portugal in the} semis.\\\\
        Ukraine\textcolor{blue}{, who} 
        won 
        \textcolor{blue}{their group} ...
        }\\
		\hlinewd{0.75pt}
	\end{tabular}}
    \caption{One case on Wikinews benchmark, and all annotators consistently judge that MD's generation is better. The prefix is highlighted in bold with an underline. Both contrastive search and momentum decoding generate very high-quality and fluent generations. However, the generation of contrastive search is slightly incoherent with the given prefix, while momentum decoding doesn't.
    As for the generation of momentum decoding,
    the new tokens outside its current directed graph are highlighted in \textcolor{red}{red}. The tokens that exist in the directed graph but still have the highest scores after the Eq. (\ref{eq:momentum_decoding})'s modification are highlighted in \textcolor{blue}{blue}.
    }
	\label{tb:case_examples}
\end{table*}

\section{Further Analysis}
\label{sec:analysis}

In this section, we provide three in-depth analyses to reveal the merits of  momentum decoding in detail:
(1) the connection with the state-of-the-art decoding method, contrastive search;
(2) comprehensive comparison between MD and baselines;
(3) the ablation study of the resistance function.

\subsection{Connection with Contrastive Search}

The formulation of contrastive search is shown in Eq. (\ref{eq:contrastive_search}). At $i$-th decoding step, contrastive search collects top-$k$ candidate tokens $V^{(k)}$ and feeds them into LMs again to obtain the hidden states $h_v$, which is used to compute their degeneration penalties (maximum cosine similarity with $\boldsymbol{x}_{<i}$).
\begin{equation}
    \label{eq:contrastive_search}
    \begin{split}
    x &= \argmax_{v\in V^{(k)}}\bigg\{(1 - \alpha)\times \underbrace{p_{\theta}(v|\boldsymbol{x}_{<i})}_{\textup{model confidence}} -  \\&\: \alpha \times \underbrace{(\max\{s(h_v, h_{x_j}):1\leq j \leq i-1\})}_{\textup{degeneration penalty}}\bigg\}
    \end{split}
\end{equation}
Through comparing it with Eq. (\ref{eq:momentum_decoding}), we could make two conclusions:
(1) the term degeneration penalty in contrastive search can be viewed as an implementation of the resistance function in momentum decoding. In our paper, the resistance function is a simple yet empirically effective look-up table, leading to better inference efficiency than the contrastive search;
(2) momentum decoding only modifies the probabilities of candidates when the top-$1$ token exists in $G$. On the contrary, contrastive search modifies the probabilities at every decoding step, leading to a higher gap between the training and inference stage. For example, as shown in Table \ref{tb:main_results}, the greedy ratio of momentum decoding is higher than the contrastive search on three benchmarks.



\begin{figure}[t]     
  \center{\includegraphics[width=0.48\textwidth] {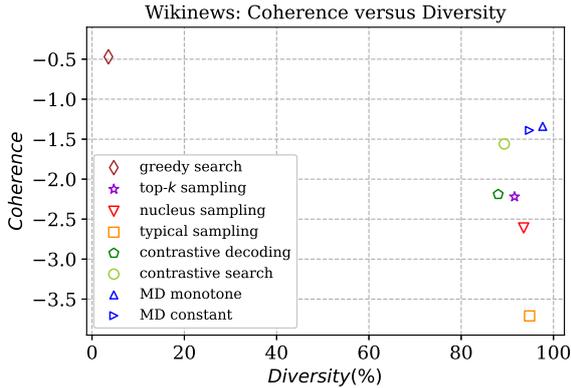}}
  \caption{Ablation study of resistance function on Wikinews benchmark. MD monotone denotes the monotone increasing resistance function designed in Table \ref{tb:resistance_function}. MD constant indicates the constant resistance function (i.e. $f(\cdot)=2$) for momentum decoding.}
  \label{img:rf_function}
\end{figure}

\subsection{Momentum Decoding versus Previous Works}
\begin{figure}[t]     
  \center{\includegraphics[width=0.48\textwidth] {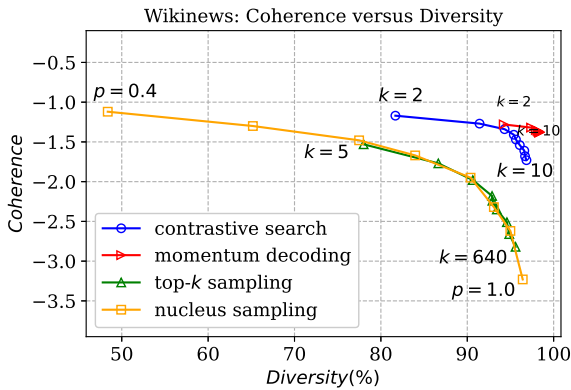}}        
    \caption{Diversity-Coherence analysis on the Wikinews benchmark.}
  \label{img:diversity_vs_coherence}
\end{figure}

We vary the hyper-parameters for different methods, i.e., $k$ for top-$k$ sampling (from 5 to 640); $p$ for nucleus
sampling (from 0.4 to 1.0); $k$ for contrastive search (from 2 to 10); and $k$ for momentum decoding (from 2 to 10)\footnote{(i) For top-$k$ sampling, $k\in [5, 10, 20, 40, 50, 80, 160, 320,$ $640]$; (ii) for nucleus sampling, $p\in
[0.4, 0.5, 0.6, 0.7, 0.8$, $0.9, 0.95, 1.0]$; (iii) for contrastive search, $k\in [2, 3, 4, 5, 6$, $7, 8, 9, 10]$; and (iv) for momentum decoding,$k\in [2, 3, 4, 5,$ $6, 7, 8, 9, 10]$. We keep $\alpha$ for contrastive search and momentum decoding as a constant 0.6 and 0.2, respectively.}.
As shown in Figure \ref{img:diversity_vs_coherence}, it can be found that our proposed momentum decoding (\textcolor{red}{red} line) notably outperforms other baselines on balance between the coherence and diversity metrics.
Besides, the gap between contrastive search and momentum decoding is relatively small, which is highly correlated with human judgments in Section (\cref{sec:human_evaluation}).
This observation probably marks the higher correlation between human judgments and the diversity-coherence combination.

\subsection{Ablation Study of Resistance Functions}

In this section, we conduct the ablation study of the resistance function. Specifically, we simply replace the monotone increasing resistance function as designed in Table \ref{tb:resistance_function} with a constant function (i.e. $f(\cdot)=2$), and the $\alpha$ hyper-parameter keeps the same $0.2$.
From Figure \ref{img:rf_function}, it can be found that the resistance function's implementation slightly influences the momentum decoding performance. Even if $f(\cdot)$ is a constant function, it could still generate text with higher diversity and coherence than contrastive search, proving the robustness of our proposed momentum decoding.

\section{Conclusion and Future Work}
In this paper, we introduce a new perspective on the task of open-ended text generation. Specifically, we view it as an exploration process within a directed graph. We understand the degeneration problem as circular loops in the directed graph. Furthermore, we propose a novel decoding method---{momentum decoding}---which encourages the LM to greedily explore the new tokens outside the current directed graph. Meanwhile, it also allows the LM to return to the existing nodes but with a momentum downgraded by a simple yet effective resistance function. We extensively test our approach on three benchmarks across different domains. Both automatic and human evaluations verify that momentum decoding performs comparably with the current state of the art while enjoying 30\% of inference speedup and more than 4$\times$ reduction in computation FLOPs.


We note that momentum decoding is model architecture-agnostic and can be applied to any generative model. In future work, we would like to extend our investigation on momentum decoding to other generative models (e.g., encoder-decoder models) and other generation tasks (e.g., machine translation and document summarization).

\section*{Limitations}
While momentum decoding achieves impressive inference efficiency and effectiveness, the current design of the resistance function (Table~\ref{tb:resistance_function}) inevitably leads to the \textit{N-gram blocking} problem. 
It can be found that, given the context $\boldsymbol{x}_{<t}$, a candidate token $x_t$ with circular depth $d(x_t,\boldsymbol{x}_{<t})\geq 4$ 
(see Eq. (\ref{eq:depth_function})) would receive a $5\times 0.2=1$ resistance in Eq. (\ref{eq:momentum_decoding}). In such case, $x_t$ with circular depth $d(x_t,\boldsymbol{x}_{<t})\geq 4$ will be blocked during generation. While as demonstrated in Appendix \ref{appendix:n_gram_stat}, the repetitions of human-written text in $n$-gram, where $n\geq4$, are reasonably low, such N-gram blocking behavior is still undesirable.

Therefore, we point out two potential solutions for this problem:
(1) adopting a smaller $\alpha$ parameter in Eq. (\ref{eq:momentum_decoding}) or carefully defining the resistance function to allow the LM to generate longer N-grams;
(2) purposely deleting earlier tokens or edges in the directed graph to lower the circular depth, allowing longer N-grams to be generated.

\bibliography{anthology,custom}

\begin{thebibliography}{25}
\expandafter\ifx\csname natexlab\endcsname\relax\def\natexlab#1{#1}\fi

\bibitem[{Basu et~al.(2020)Basu, Ramachandran, Keskar, and
  Varshney}]{basu2020mirostat}
Sourya Basu, Govardana~Sachitanandam Ramachandran, Nitish~Shirish Keskar, and
  Lav~R Varshney. 2020.
\newblock Mirostat: A neural text decoding algorithm that directly controls
  perplexity.
\newblock \emph{arXiv preprint arXiv:2007.14966}.

\bibitem[{Basu et~al.(2021)Basu, Ramachandran, Keskar, and
  Varshney}]{Basu2021MirostatAN}
Sourya Basu, Govardana~Sachithanandam Ramachandran, Nitish~Shirish Keskar, and
  Lav~R. Varshney. 2021.
\newblock Mirostat: a neural text decoding algorithm that directly controls
  perplexity.
\newblock In \emph{International Conference on Learning Representations}.

\bibitem[{Fan et~al.(2018{\natexlab{a}})Fan, Lewis, and
  Dauphin}]{fan2018hierarchical}
Angela Fan, Mike Lewis, and Yann Dauphin. 2018{\natexlab{a}}.
\newblock Hierarchical neural story generation.
\newblock \emph{arXiv preprint arXiv:1805.04833}.

\bibitem[{Fan et~al.(2018{\natexlab{b}})Fan, Lewis, and
  Dauphin}]{Fan2018HierarchicalNS}
Angela Fan, Mike Lewis, and Yann Dauphin. 2018{\natexlab{b}}.
\newblock Hierarchical neural story generation.
\newblock In \emph{Annual Meeting of the Association for Computational
  Linguistics}.

\bibitem[{Holtzman et~al.(2020)Holtzman, Buys, Forbes, and
  Choi}]{Holtzman2020TheCC}
Ari Holtzman, Jan Buys, Maxwell Forbes, and Yejin Choi. 2020.
\newblock The curious case of neural text degeneration.
\newblock \emph{ArXiv}, abs/1904.09751.

\bibitem[{Li et~al.(2016)Li, Galley, Brockett, Gao, and Dolan}]{Li2016ADO}
Jiwei Li, Michel Galley, Chris Brockett, Jianfeng Gao, and William~B. Dolan.
  2016.
\newblock A diversity-promoting objective function for neural conversation
  models.
\newblock In \emph{North American Chapter of the Association for Computational
  Linguistics}.

\bibitem[{Li et~al.(2022)Li, Holtzman, Fried, Liang, Eisner, Hashimoto,
  Zettlemoyer, and Lewis}]{Li2022ContrastiveDO}
Xiang~Lisa Li, Ari Holtzman, Daniel Fried, Percy Liang, Jason Eisner, Tatsunori
  Hashimoto, Luke Zettlemoyer, and Mike Lewis. 2022.
\newblock Contrastive decoding: Open-ended text generation as optimization.
\newblock \emph{ArXiv}, abs/2210.15097.

\bibitem[{Liu et~al.(2020)Liu, Zhou, Zhao, Wang, Deng, and
  Ju}]{Liu2020FastBERTAS}
Weijie Liu, Peng Zhou, Zhe Zhao, Zhiruo Wang, Haotang Deng, and Qi~Ju. 2020.
\newblock Fastbert: a self-distilling bert with adaptive inference time.
\newblock In \emph{Annual Meeting of the Association for Computational
  Linguistics}.

\bibitem[{Meister et~al.(2022)Meister, Pimentel, Wiher, and
  Cotterell}]{Meister2022TypicalDF}
Clara Meister, Tiago Pimentel, Gian Wiher, and Ryan Cotterell. 2022.
\newblock Typical decoding for natural language generation.
\newblock \emph{ArXiv}, abs/2202.00666.

\bibitem[{Merity et~al.(2017)Merity, Xiong, Bradbury, and
  Socher}]{Merity2017PointerSM}
Stephen Merity, Caiming Xiong, James Bradbury, and Richard Socher. 2017.
\newblock Pointer sentinel mixture models.
\newblock \emph{ArXiv}, abs/1609.07843.

\bibitem[{Mostafazadeh et~al.(2016)Mostafazadeh, Chambers, He, Parikh, Batra,
  Vanderwende, Kohli, and Allen}]{Mostafazadeh2016ACA}
N.~Mostafazadeh, Nathanael Chambers, Xiaodong He, Devi Parikh, Dhruv Batra,
  Lucy Vanderwende, Pushmeet Kohli, and James~F. Allen. 2016.
\newblock A corpus and cloze evaluation for deeper understanding of commonsense
  stories.
\newblock In \emph{North American Chapter of the Association for Computational
  Linguistics}.

\bibitem[{Pillutla et~al.(2021)Pillutla, Swayamdipta, Zellers, Thickstun,
  Welleck, Choi, and Harchaoui}]{Pillutla2021MAUVEMT}
Krishna Pillutla, Swabha Swayamdipta, Rowan Zellers, John Thickstun, Sean
  Welleck, Yejin Choi, and Za{\"i}d Harchaoui. 2021.
\newblock Mauve: Measuring the gap between neural text and human text using
  divergence frontiers.
\newblock In \emph{Neural Information Processing Systems}.

\bibitem[{Radford et~al.(2019)Radford, Wu, Child, Luan, Amodei, and
  Sutskever}]{Radford2019LanguageMA}
Alec Radford, Jeff Wu, Rewon Child, David Luan, Dario Amodei, and Ilya
  Sutskever. 2019.
\newblock Language models are unsupervised multitask learners.

\bibitem[{Rae et~al.(2021)Rae, Borgeaud, Cai, Millican, Hoffmann, Song,
  Aslanides, Henderson, Ring, Young, Rutherford, Hennigan, Menick, Cassirer,
  Powell, van~den Driessche, Hendricks, Rauh, Huang, Glaese, Welbl, Dathathri,
  Huang, Uesato, Mellor, Higgins, Creswell, McAleese, Wu, Elsen, Jayakumar,
  Buchatskaya, Budden, Sutherland, Simonyan, Paganini, Sifre, Martens, Li,
  Kuncoro, Nematzadeh, Gribovskaya, Donato, Lazaridou, Mensch, Lespiau,
  Tsimpoukelli, Grigorev, Fritz, Sottiaux, Pajarskas, Pohlen, Gong, Toyama,
  de~Masson~d'Autume, Li, Terzi, Mikulik, Babuschkin, Clark, de~Las~Casas, Guy,
  Jones, Bradbury, Johnson, Hechtman, Weidinger, Gabriel, Isaac, Lockhart,
  Osindero, Rimell, Dyer, Vinyals, Ayoub, Stanway, Bennett, Hassabis,
  Kavukcuoglu, and Irving}]{Rae2021ScalingLM}
Jack~W. Rae, Sebastian Borgeaud, Trevor Cai, Katie Millican, Jordan Hoffmann,
  Francis Song, John Aslanides, Sarah Henderson, Roman Ring, Susannah Young,
  Eliza Rutherford, Tom Hennigan, Jacob Menick, Albin Cassirer, Richard Powell,
  George van~den Driessche, Lisa~Anne Hendricks, Maribeth Rauh, Po-Sen Huang,
  Amelia Glaese, Johannes Welbl, Sumanth Dathathri, Saffron Huang, Jonathan
  Uesato, John F.~J. Mellor, Irina Higgins, Antonia Creswell, Nathan McAleese,
  Amy Wu, Erich Elsen, Siddhant~M. Jayakumar, Elena Buchatskaya, David Budden,
  Esme Sutherland, Karen Simonyan, Michela Paganini, L.~Sifre, Lena Martens,
  Xiang~Lorraine Li, Adhiguna Kuncoro, Aida Nematzadeh, Elena Gribovskaya,
  Domenic Donato, Angeliki Lazaridou, Arthur Mensch, Jean-Baptiste Lespiau,
  Maria Tsimpoukelli, N.~K. Grigorev, Doug Fritz, Thibault Sottiaux, Mantas
  Pajarskas, Tobias Pohlen, Zhitao Gong, Daniel Toyama, Cyprien
  de~Masson~d'Autume, Yujia Li, Tayfun Terzi, Vladimir Mikulik, Igor
  Babuschkin, Aidan Clark, Diego de~Las~Casas, Aurelia Guy, Chris Jones, James
  Bradbury, Matthew~G. Johnson, Blake~A. Hechtman, Laura Weidinger, Iason
  Gabriel, William~S. Isaac, Edward Lockhart, Simon Osindero, Laura Rimell,
  Chris Dyer, Oriol Vinyals, Kareem~W. Ayoub, Jeff Stanway, L.~L. Bennett,
  Demis Hassabis, Koray Kavukcuoglu, and Geoffrey Irving. 2021.
\newblock Scaling language models: Methods, analysis \& insights from training
  gopher.
\newblock \emph{ArXiv}, abs/2112.11446.

\bibitem[{Su et~al.(2021{\natexlab{a}})Su, Cai, Zhou, Lin, Baker, Cao, Shi,
  Collier, and Wang}]{su2021dialogue}
Yixuan Su, Deng Cai, Qingyu Zhou, Zibo Lin, Simon Baker, Yunbo Cao, Shuming
  Shi, Nigel Collier, and Yan Wang. 2021{\natexlab{a}}.
\newblock Dialogue response selection with hierarchical curriculum learning.
\newblock In \emph{Proceedings of the 59th Annual Meeting of the Association
  for Computational Linguistics and the 11th International Joint Conference on
  Natural Language Processing (Volume 1: Long Papers)}, pages 1740--1751.

\bibitem[{Su and Collier(2022)}]{su2022contrastiveiswhatyouneed}
Yixuan Su and Nigel Collier. 2022.
\newblock Contrastive search is what you need for neural text generation.
\newblock \emph{arXiv preprint arXiv:2210.14140}.

\bibitem[{Su et~al.(2022{\natexlab{a}})Su, Lan, Liu, Liu, Yogatama, Wang, Kong,
  and Collier}]{su2022language}
Yixuan Su, Tian Lan, Yahui Liu, Fangyu Liu, Dani Yogatama, Yan Wang, Lingpeng
  Kong, and Nigel Collier. 2022{\natexlab{a}}.
\newblock Language models can see: Plugging visual controls in text generation.
\newblock \emph{arXiv preprint arXiv:2205.02655}.

\bibitem[{Su et~al.(2022{\natexlab{b}})Su, Lan, Wang, Yogatama, Kong, and
  Collier}]{Su2022ACF}
Yixuan Su, Tian Lan, Yan Wang, Dani Yogatama, Lingpeng Kong, and Nigel Collier.
  2022{\natexlab{b}}.
\newblock \href {https://openreview.net/forum?id=V88BafmH9Pj} {A contrastive
  framework for neural text generation}.
\newblock In \emph{Advances in Neural Information Processing Systems}.

\bibitem[{Su et~al.(2022{\natexlab{c}})Su, Shu, Mansimov, Gupta, Cai, Lai, and
  Zhang}]{Su2022MultiTaskPF}
Yixuan Su, Lei Shu, Elman Mansimov, Arshit Gupta, Deng Cai, Yi-An Lai, and
  Yi~Zhang. 2022{\natexlab{c}}.
\newblock Multi-task pre-training for plug-and-play task-oriented dialogue
  system.
\newblock In \emph{Annual Meeting of the Association for Computational
  Linguistics}.

\bibitem[{Su et~al.(2021{\natexlab{b}})Su, Wang, Baker, Cai, Liu, Korhonen, and
  Collier}]{Su2021PROTOTYPETOSTYLEDG}
Yixuan Su, Yan Wang, Simon Baker, Deng Cai, Xiaojiang Liu, Anna Korhonen, and
  Nigel Collier. 2021{\natexlab{b}}.
\newblock Prototype-to-style: Dialogue generation with style-aware editing on
  retrieval memory.
\newblock \emph{IEEE/ACM Transactions on Audio, Speech, and Language
  Processing}, 29:2152--2161.

\bibitem[{Su and Xu(2022)}]{su2022empirical}
Yixuan Su and Jialu Xu. 2022.
\newblock An empirical study on contrastive search and contrastive decoding for
  open-ended text generation.
\newblock \emph{arXiv preprint arXiv:2211.10797}.

\bibitem[{Thoppilan et~al.(2022)Thoppilan, Freitas, Hall, Shazeer,
  Kulshreshtha, Cheng, Jin, Bos, Baker, Du, Li, Lee, Zheng, Ghafouri, Menegali,
  Huang, Krikun, Lepikhin, Qin, Chen, Xu, Chen, Roberts, Bosma, Zhou, Chang,
  Krivokon, Rusch, Pickett, Meier-Hellstern, Morris, Doshi, Santos, Duke,
  S{\o}raker, Zevenbergen, Prabhakaran, D{\'i}az, Hutchinson, Olson, Molina,
  Hoffman-John, Lee, Aroyo, Rajakumar, Butryna, Lamm, Kuzmina, Fenton, Cohen,
  Bernstein, Kurzweil, Aguera-Arcas, Cui, Croak, Chi, and
  Le}]{Thoppilan2022LaMDALM}
Romal Thoppilan, Daniel~De Freitas, Jamie Hall, Noam~M. Shazeer, Apoorv
  Kulshreshtha, Heng-Tze Cheng, Alicia Jin, Taylor Bos, Leslie Baker, Yu~Du,
  Yaguang Li, Hongrae Lee, Huaixiu Zheng, Amin Ghafouri, Marcelo Menegali,
  Yanping Huang, Maxim Krikun, Dmitry Lepikhin, James Qin, Dehao Chen,
  Yuanzhong Xu, Zhifeng Chen, Adam Roberts, Maarten Bosma, Yanqi Zhou,
  Chung-Ching Chang, I.~A. Krivokon, Willard~James Rusch, Marc Pickett,
  Kathleen~S. Meier-Hellstern, Meredith~Ringel Morris, Tulsee Doshi,
  Renelito~Delos Santos, Toju Duke, Johnny~Hartz S{\o}raker, Ben Zevenbergen,
  Vinodkumar Prabhakaran, Mark D{\'i}az, Ben Hutchinson, Kristen Olson,
  Alejandra Molina, Erin Hoffman-John, Josh Lee, Lora Aroyo, Ravindran
  Rajakumar, Alena Butryna, Matthew Lamm, V.~O. Kuzmina, Joseph Fenton, Aaron
  Cohen, Rachel Bernstein, Ray Kurzweil, Blaise Aguera-Arcas, Claire Cui,
  Marian Croak, Ed~Chi, and Quoc Le. 2022.
\newblock Lamda: Language models for dialog applications.
\newblock \emph{ArXiv}, abs/2201.08239.

\bibitem[{Zhang et~al.(2022)Zhang, Roller, Goyal, Artetxe, Chen, Chen, Dewan,
  Diab, Li, Lin, Mihaylov, Ott, Shleifer, Shuster, Simig, Koura, Sridhar, Wang,
  and Zettlemoyer}]{Zhang2022OPTOP}
Susan Zhang, Stephen Roller, Naman Goyal, Mikel Artetxe, Moya Chen, Shuohui
  Chen, Christopher Dewan, Mona Diab, Xian Li, Xi~Victoria Lin, Todor Mihaylov,
  Myle Ott, Sam Shleifer, Kurt Shuster, Daniel Simig, Punit~Singh Koura, Anjali
  Sridhar, Tianlu Wang, and Luke Zettlemoyer. 2022.
\newblock Opt: Open pre-trained transformer language models.
\newblock \emph{ArXiv}, abs/2205.01068.

\bibitem[{Zhang et~al.(2019)Zhang, Feng, Meng, You, and
  Liu}]{zhang2019bridging}
Wen Zhang, Yang Feng, Fandong Meng, Di~You, and Qun Liu. 2019.
\newblock Bridging the gap between training and inference for neural machine
  translation.
\newblock \emph{arXiv preprint arXiv:1906.02448}.

\bibitem[{Zhu et~al.(2015)Zhu, Kiros, Zemel, Salakhutdinov, Urtasun, Torralba,
  and Fidler}]{Zhu2015AligningBA}
Yukun Zhu, Ryan Kiros, Richard~S. Zemel, Ruslan Salakhutdinov, Raquel Urtasun,
  Antonio Torralba, and Sanja Fidler. 2015.
\newblock Aligning books and movies: Towards story-like visual explanations by
  watching movies and reading books.
\newblock \emph{2015 IEEE International Conference on Computer Vision (ICCV)},
  pages 19--27.

\end{thebibliography}
\bibliographystyle{acl_natbib}

\clearpage
\appendix

\section{Pseudo-code of Momentum Decoding}
\label{appendix:alg}
Algorithm \ref{alg:code} provides the pseudo-code of our proposed momentum decoding. At each decoding step, momentum decoding considers ranking top-$k$ candidate tokens ($k=5$ in this study). The greedy search is conducted if the top-$1$ candidate token doesn't exist in the currently directed graph. Otherwise, the scores of top-$k$ candidate tokens are calculated by Eq. (\ref{eq:momentum_decoding}), and the one with the highest score is chosen as the next token.
Since the computation cost of updating and searching $G$ is neglectable, the inference efficiency of momentum decoding is significantly closer to the greedy search.

\begin{algorithm}[h]
\caption{Pseudo-code of Momentum Decoding in a PyTorch-like style.}
\label{alg:code}
\algcomment{\fontsize{7.2pt}{0em}\selectfont \texttt{softmax}: softmax activation function; \texttt{argmax}: argmax function;
\texttt{topk}: topk function.
}
\definecolor{codeblue}{rgb}{0.25,0.5,0.5}
\lstset{
  backgroundcolor=\color{white},
  basicstyle=\fontsize{7.2pt}{7.2pt}\ttfamily\selectfont,
  columns=fullflexible,
  breaklines=true,
  captionpos=b,
  commentstyle=\fontsize{7.2pt}{7.2pt}\color{codeblue},
  keywordstyle=\fontsize{7.2pt}{7.2pt},
}
\begin{lstlisting}[language=python]
# LM: language model, for example, the GPT2
# init_graph: function to initialize the directed graph for the given prefix
# prefix: a sequence of tokens in prefix
# decoding_length: the number of the new tokens that should be generated
# top_k: top-k candidate tokens
# find_depth: function to find depths of candidate token in directed graph

dg = init_graph(prefix)  # initialized the directed graph
for _ in range(decoding_length):  # auto-regressive decoding
    # logits over the vocabulary
    logits = LM(prefix)
    # probability of top-k candidate tokens
    top_k_ids, top_k_probs = softmax(logits, dim=-1).topk(k=top_k)
    
    if top_k_ids[0] not in dg:  
        # top-1 token doesn't in directed graph, conduct greedy search
        next_token = top_k_ids[0]
    else:  
        # calculate the penalty by the resistance function
        depth = [max(find_depth(dg,x)) for x in top_k_ids]
        resistance = [RF(d) for d in depth]
        scores = [p - alpha * r for p, r in zip(top_k_probs, resistance)]
        next_token = argmax(scores)
        
    # update the directed graph
    dg.update(next_token)
    
    # update the prefix
    prefix.append(next_token)
\end{lstlisting}
\end{algorithm}

\begin{table}[t]
\small
	\centering  
	\renewcommand{\arraystretch}{1.2}
	\setlength{\tabcolsep}{6pt}
	\scalebox{0.88}{
        \begin{tabular}{cccc}
\hlinewd{0.75pt}
\textbf{$N$-gram Repetitions} & \textbf{Wikinews} & \textbf{Wikitext} & \textbf{BookCorpus} \\ \hlinewd{0.75pt}
\textbf{2-gram}           &  10.76\%                 &  9.47\%                 &  7.14\%                   \\
\textbf{3-gram}           &   2.49\%                &      2.94\%             &    1.51\%                 \\
\textbf{4-gram}           &   0.86\%                &      1.05\%             &     0.48\%                \\
\textbf{5-gram}           &  0.38\%                 &      0.46\%             &     0.19\%                \\
\textbf{6-gram}           &  0.19\%                 &      0.22\%             &    0.09\%                 \\
\textbf{7-gram}           &  0.09\%                 &      0.11\%             &   0.05\%                  \\ 
\textbf{8-gram}           &  0.06\%                 &      0.07\%             &    0.04\%                 \\\hlinewd{0.75pt}
\end{tabular}
    }
    \caption{The proportion of the repetition $n$-grams in three benchmarks. It can be found that the proportion of the repetition $4$-gram is extremely slow.}
\label{tb:ngram_repetition_stat}
\end{table}

\section{More Cases}

\begin{table*}[t]
    \small
	\centering  
	\renewcommand{\arraystretch}{1.2}
	\setlength{\tabcolsep}{6pt}
	\scalebox{0.76}{
	\begin{tabular}{cc}
	    \hlinewd{0.75pt}
        \textbf{Method}&\textbf{Generated Result}\\
        \hline
        \makecell[c]{Contrastive\\  Search}&\makecell[l]{\underline{\textbf{At least ten people are dead after a large, destructive tornado tore through Yazoo City, Mississippi yesterday. The tornado has been rated}}\\ \underline{\textbf{an EF4 on the Enhanced}} Fujita Scale    , which is used to measure tornadoes in the U.S. and Canada.\\\\Here's a look at the damage in Yazoo City, courtesy of WLOX-TV in Biloxi, Mississippi:\\\\Advertisement. Article Continues Below. Video by Chris Cappella/Shutterstock.\\\\According to Weather Underground, the twister touched down at 2:30 pm local time (11:30 am ET) near the intersection of I-10 and U.S. Highway 61.\\ As of this writing, there are reports of at least ten fatalities, including a woman who was killed by a falling tree.\\\\Advertisement. Article Continues Below. Video by Chris Cappella/Shutterstock.\\\\The National Weather Service is warning that a ``large number of power outages are expected to continue for the next 24 hours, as crews work to\\ restore power to as many customers as possible.'' In addition to the storm damage, the city is dealing with water and sewage overflows, which is\\ a common problem in this time of year.\\\\Advertisement. Article Continues Below. Video by Chris Cappella/Shutterstock.\\\\The Associated Press reports that some schools have been closed for the day, ...}\\
        \hline
        \makecell[c]{Momentum\\ Decoding}&\makecell[l]{\underline{\textbf{At least ten people are dead after a large, destructive tornado tore through Yazoo City, Mississippi yesterday. The tornado has been rated}}\\ \underline{\textbf{an EF4 on the Enhanced}}
        \textcolor{red}{Fujita Scale}\textcolor{blue}{,} 
        \textcolor{red}{which is} used 
        \textcolor{red}{to measure tornadoes}\textcolor{blue}{.}\\\\
        \textcolor{red}{The} 
        \textcolor{blue}{tornado} 
        \textcolor{red}{was reported at around 2:30 p}\textcolor{blue}{.}\textcolor{red}{m}\textcolor{blue}{.} \textcolor{red}{and} touched 
        \textcolor{red}{down in} \textcolor{blue}{the} 
        \textcolor{red}{area of} Highway \textcolor{red}{59} \textcolor{blue}{and} Interstate \textcolor{red}{24}\textcolor{blue}{. It was} 
        \textcolor{red}{moving west}-\textcolor{red}{northwest} 
        \textcolor{blue}{at} about\\ 
        \textcolor{red}{50 miles per hour}\textcolor{blue}{.} According \textcolor{blue}{to the} 
        \textcolor{red}{National Weather Service}
        \textcolor{blue}{, the tornado} 
        had \textcolor{red}{winds} 
        \textcolor{blue}{of} 
        \textcolor{red}{up} 
        \textcolor{blue}{to} 
        \textcolor{red}{100 mph}\textcolor{blue}{.}\\\\
        ``\textcolor{red}{It}'s just devastating\textcolor{red}{,'' said} one \textcolor{red}{resident who lives near} \textcolor{blue}{the} 
        scene 
        \textcolor{red}{of the} 
        storm\textcolor{blue}{.} 
        \textcolor{red}{``I've never seen anything like it.``}\\\\
        According \textcolor{blue}{to} \textcolor{red}{WREG}\textcolor{blue}{,} 
        there \textcolor{red}{were no reports} 
        \textcolor{blue}{of} 
        \textcolor{red}{injuries} or \textcolor{red}{fatalities}\textcolor{blue}{.} 
        A \textcolor{red}{number} 
        \textcolor{blue}{of} 
        \textcolor{red}{homes} 
        \textcolor{blue}{were} 
        \textcolor{red}{damaged} 
        by 
        \textcolor{blue}{the} powerful 
        \textcolor{blue}{tornado.}\\\\
        A witness \textcolor{red}{told} \textcolor{blue}{WREG} \textcolor{red}{that} 
        he 
        \textcolor{red}{saw} 
        \textcolor{blue}{a} 
        \textcolor{red}{car go} 
        \textcolor{blue}{through a} 
        \textcolor{red}{fence} 
        \textcolor{blue}{and} 
        \textcolor{red}{into} 
        \textcolor{blue}{a} 
        \textcolor{red}{home}\textcolor{blue}{.} 
        He also \textcolor{blue}{said that} his 
        \textcolor{red}{house} 
        \textcolor{blue}{was} 
        destroyed\textcolor{blue}{.}\\\\
        Another \textcolor{blue}{witness} said she \textcolor{blue}{saw a} man get \textcolor{red}{hit} \textcolor{blue}{by} debris from \textcolor{blue}{the tornado.} 
        She \textcolor{blue}{said} the \textcolor{blue}{man was} \textcolor{red}{taken} \textcolor{blue}{to} a \textcolor{red}{hospital} with \textcolor{red}{non}\textcolor{blue}{-}\textcolor{red}{life threatening} \textcolor{blue}{injuries.}\\\\
        According \textcolor{blue}{to} \textcolor{red}{The} \textcolor{red}{Clarion}\textcolor{blue}{-}\textcolor{red}{Ledger}\textcolor{blue}{,} \textcolor{red}{Yazoo County Emergency Management Director Mike Boudreaux} \textcolor{blue}{said} there have \textcolor{blue}{been no} confirmed \textcolor{blue}{reports of} \\deaths \textcolor{blue}{or injuries} in the \textcolor{red}{county}\textcolor{blue}{.}\\\\\textcolor{blue}{Boudreaux} added \textcolor{blue}{that the} damage \textcolor{blue}{is} \textcolor{red}{extensive} \textcolor{blue}{and} will \textcolor{red}{take some time} \textcolor{blue}{to} \textcolor{red}{assess} ...
        }\\
		\hlinewd{0.75pt}
	\end{tabular}
    }
    \caption{One case on Wikinews benchmark, and all annotators consistently judge that MD's generation is better than the one generated by the contrastive search. The prefix is highlighted in bold with an underline. It can be found that the contrastive search also leads to invalid repetitions in this case.
    As for the generation of momentum decoding,
    the new tokens outside its current directed graph are highlighted in \textcolor{red}{red}. The tokens that exist in the directed graph but still have the highest scores after the Eq. (\ref{eq:momentum_decoding})'s modification are highlighted in \textcolor{blue}{blue}.
    }
	\label{tb:case_examples_1}
\end{table*}

\section{More Analysis}
\subsection{Ablation Study of Resistance Function}

\begin{figure*}[t]     
  \center{\includegraphics[width=1.0\textwidth] {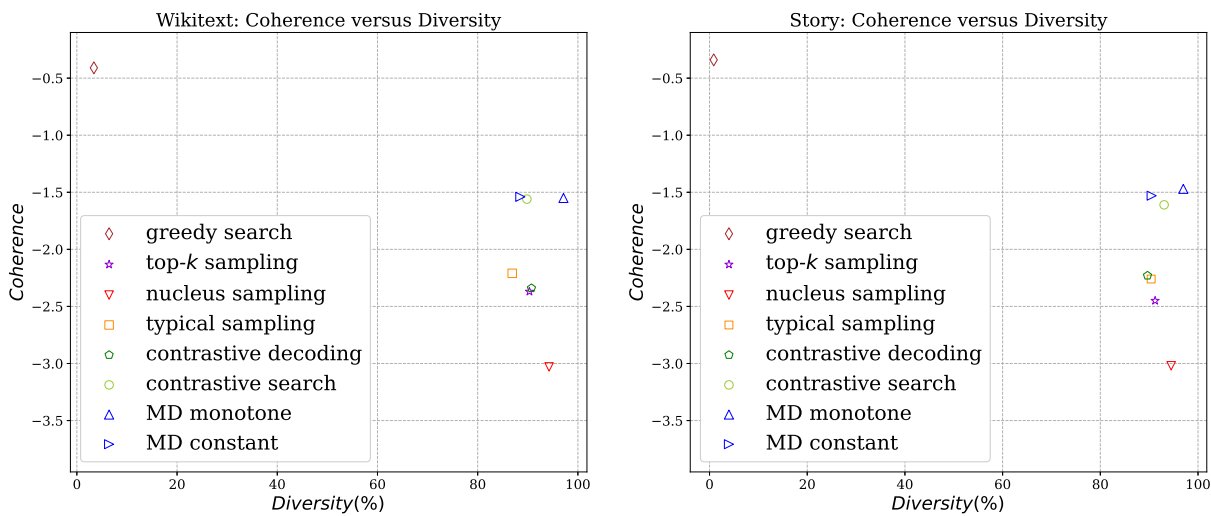}}        
  \caption{Ablation study of resistance function on Wikitext and BookCorpus (Story) benchmark. MD monotone denotes the monotone increasing resistance function designed in Table \ref{tb:resistance_function}. MD constant indicates the constant resistance function (constant is 2) for momentum decoding.}
  \label{img:diversity_vs_coherence_wikitext_story}
\end{figure*}

More ablation studies on Wikitext and BookCorpus benchmarks are shown in Figure \ref{img:diversity_vs_coherence_wikitext_story}.
It can be found that the implementation of the resistance function slightly influences the performance of momentum decoding. Even if $f(\cdot)$ is a constant function, it could still generate text with higher diversity and coherence than contrastive search, proving the robustness of our proposed momentum decoding.

\subsection{Momentum Decoding versus Previous Works}

\begin{figure*}[t]     
  \center{\includegraphics[width=1.0\textwidth] {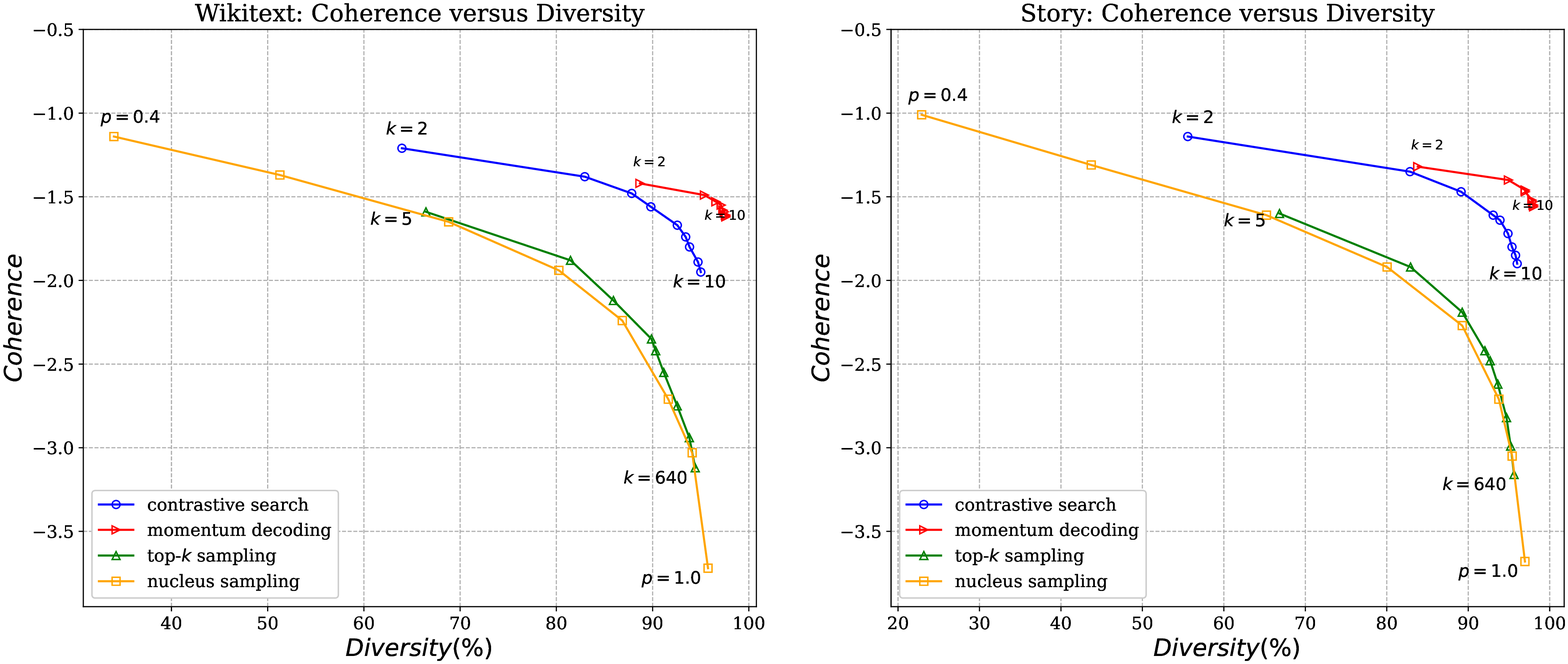}}        
  \caption{Diversity-Coherence analysis on Wikitext and Story benchmarks.}
  \label{img:wikitext_story_diversity_vs_coherence}
\end{figure*}

Figure \ref{img:wikitext_story_diversity_vs_coherence} shows the diversity-coherence balance analysis on Wikitext and BookCorpus (Story) benchmarks. It can be found that our proposed momentum decoding notably outperforms the previous baselines on balance between the coherence and diversity metrics, indicating that momentum decoding solves the degeneration problem and could generate robust and diverse text.


\section{$n$-gram Statistics on Three Benchmarks}
\label{appendix:n_gram_stat}
The repetition statistics $N$-grams are shown in Table \ref{tb:ngram_repetition_stat}. It can be found that the longer $n$-grams have an extremely slow repetition proportion in all three benchmarks.

\end{document}